\documentclass[runningheads]{llncs}
\usepackage{graphicx}
\usepackage{amssymb}
\usepackage{amsmath}
\usepackage{multirow}
\usepackage{array}
\usepackage{xcolor}
\usepackage{hyperref}
\begin{document}
\title{DeepGraphDMD: Interpretable Spatio-Temporal Decomposition of Non-linear Functional Brain Network Dynamics}
\titlerunning{DeepGraphDMD}
\author{Md Asadullah Turja\inst{1} \and Martin Styner\inst{2} \and Guorong Wu\inst{3}}
%index{Turja, Md Asadullah}
%index{Styner, Martin}
%index{Wu, Guorong}
%
\authorrunning{M.A. Turja et al.}
\institute{Department of Computer Science, University of North Carolina at Chapel Hill,
\email{mturja@cs.unc.edu}\\ \and
Department of Computer Science, University of North Carolina at Chapel Hill,
\email{styner@email.unc.edu}\\ \and
Department of Psychiatry, University of North Carolina at Chapel Hill, \email{guorong\_wu@med.unc.edu}
}
\maketitle              % typeset the header of the contribution
\begin{abstract}
Functional brain dynamics is supported by parallel and overlapping functional network modes that are associated with specific neural circuits. Decomposing these network modes from fMRI data and finding their temporal characteristics is challenging due to their time-varying nature and the non-linearity of the functional dynamics. Dynamic Mode Decomposition (DMD) algorithms have been quite popular for solving this decomposition problem in recent years. In this work, we apply GraphDMD --- an extension of the DMD for network data --- to extract the dynamic network modes and their temporal characteristics from the fMRI time series in an interpretable manner. GraphDMD, however, regards the underlying system as a linear dynamical system that is sub-optimal for extracting the network modes from non-linear functional data. In this work, we develop a generalized version of the GraphDMD algorithm --- DeepGraphDMD--- applicable to arbitrary non-linear graph dynamical systems. DeepGraphDMD is an autoencoder-based deep learning model that learns Koopman eigenfunctions for graph data and embeds the non-linear graph dynamics into a latent linear space. We show the effectiveness of our method in both simulated data and the HCP resting-state fMRI data. In the HCP data, DeepGraphDMD provides novel insights into cognitive brain functions by discovering two major network modes related to fluid and crystallized intelligence.

\end{abstract}
\section{Introduction}
The human brain has evolved to support a set of complementary and temporally varying brain network organizations enabling parallel and higher-order information processing \cite{osada,sigman2008brain,cons_long}. Decoupling these networks from a non-linear mixture of signals (such as functional MRI) and extracting their temporal characteristics in an interpretable manner has been a long-standing challenge in the neuroscience community.

Conventional mode/component decomposition methods such as Principal Component Analysis (PCA ) or Independent Component Analysis (ICA) assume the modes to be static \cite{hyvarinen1999fast,mckeown1998analysis,viviani2005functional} and thus sub-optimal for the functional networks generated by time-varying modes. Dynamic Mode Decomposition (DMD) can be treated as a dynamic extension of such component analysis methods since it allows its modes to oscillate over time with a fixed frequency \cite{schmid2010dynamic}. This assumption is appropriate for the human brain as the functional brain organizations are supported by oscillatory network modes \cite{casorso,IKEDA2022118801,dmd_rep,pmlr-v143,XIAO2022119618}. An extension of DMD for network data called GraphDMD \cite{GraphDMD} preserves the graph structure of the networks during the decomposition. In our work, we extend GraphDMD to a sequence of sliding window based dynamic functional connectivity (dNFC) networks to extract independent and oscillatory functional network modes. 

Under the hood, GraphDMD regards the network sequence as a linear dynamical system (LDS) where a linear operator shifts the current network state one time-point in the future. The LDS assumption, however, is not optimal for modeling functional brain networks that exhibit complex non-linearity such as rapid synchronization and desynchronization as well as transient events \cite{He2018RobustPerception}. Articles \cite{NIPS2008_950a4152,pmlr-v54-linderman17a} propose switching linear dynamical system (SLDS) to tackle the nonlinearity of spatiotemporal data by a piecewise linear approximation. While these models offer interpretability, their shallow architecture limits their generalizability to arbitrary nonlinear systems. On the other hand, the methods in \cite{NIPS2016_76dc611d,lfads,Krishnan2016StructuredIN} model the non-linearity with a deep neural network. While these models have more representation capabilities compared to SLDS, the latent states are not interpretable. More importantly, all of these methods consider the node-level dynamics instead of the network dynamics.

\begin{figure}[tb]
    \centering
    \includegraphics[width=\textwidth]{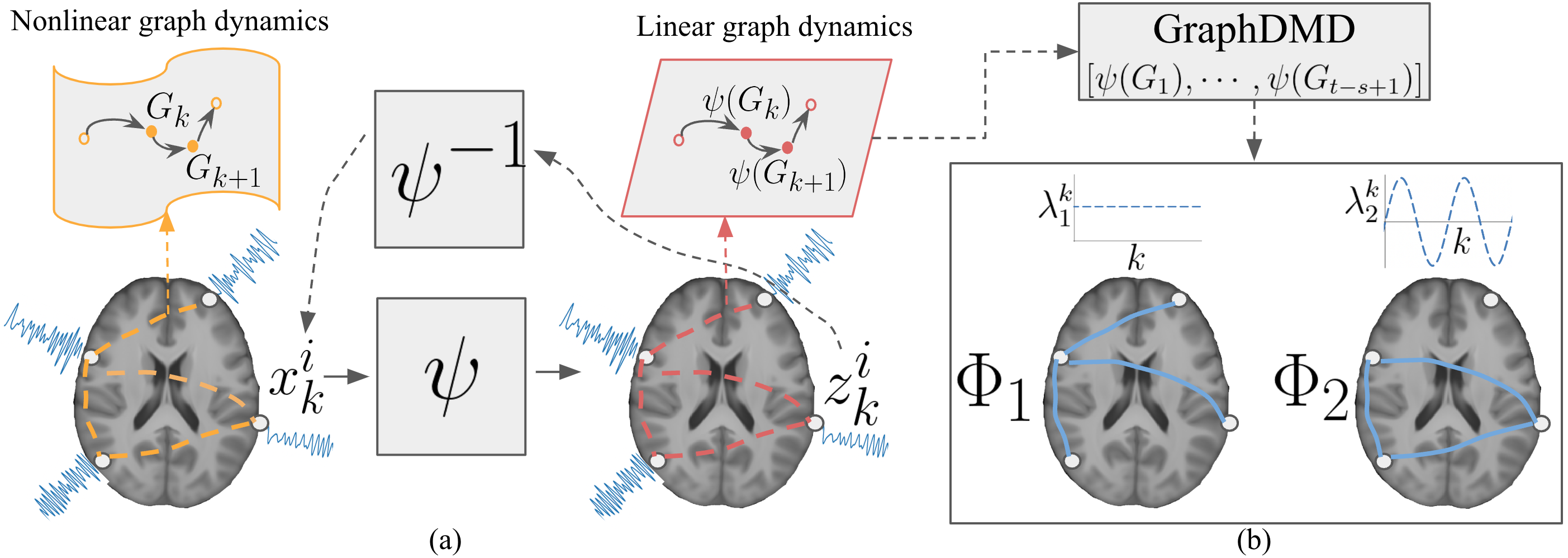}
    \caption{(a) Illustration of the DeepGraphDMD model that embeds a nonlinear graph dynamical system into a linear space, and, (b) interpretable dynamic modes and their temporal characteristics after applying GraphDMD in the linear space.}
    \label{fig:fig1}
\end{figure}

Here, we propose a novel Deep Graph Dynamic Mode Decomposition (DeepGraphDMD) algorithm that applies to arbitrary non-linear network dynamics while maintaining interpretability in the latent space. Our method uses Koopman operator theory to lift a non-linear dynamical system into a linear space through a set of Koopman eigenfunctions (Figure~\ref{fig:fig1}a). There has been a growing line of work that learns these measurement functions using deep autoencoder architectures \cite{LuschDeepDynamics,lkis}. Training these autoencoders for network data, however, has two unique challenges -- 1. preserving the edge identity in the latent space so that the network modes are interpretable, 2. enforcing linearity in the latent space for the high dimensional network data. In DeepGraphDMD, we tackle the first challenge by indirectly computing the network embeddings by a novel node embedding scheme. For the second challenge, we introduce a sparse Koopman operator to reduce the complexity of the learning problem. We evaluate the effectiveness of our novel method in both simulated data and resting-state fMRI (rs-fMRI) data from Human Connectome Project.

\section{Methodology}
Let's assume $X \in \mathbb{R}^{n \times t}$ is a matrix containing the BOLD (blood-oxygen-level-dependent) signal of $n$ brain regions (ROIs) in its rows at $t$ time frames sampled at every $k\Delta t$ time points, where $\Delta t$ is the temporal resolution. To compute the dynamic connectivity matrix at time point $k \Delta t$, a snapshot $X_k = X_{:,k:k+s}$ is taken in a sliding window of $s$ time frames. A correlation matrix $G_k \in \mathbb{R}^{n \times n}$ is then computed from $X_k$ by taking the pearson correlation between the rows of $X_k$, i.e., $G_k^{ij} = \texttt{pearson}(x_k^i, x_k^j)$ where $x_k^i$, $x_k^j$ are the $i^{th}$ and $j^{th}$ row of $X_k$ respectively. This yields a sequence of graphs $G = [G_1, G_2, \cdots, G_{t-s+1}]$. Let's also assume that $g_k \in \mathcal{R}^{n^2}$ is a vectorized version of $G_k$, i.e. $g_k = vec(G_k)$ and $g \in \mathcal{R}^{n^2 \times (t-s+1)}$ is a matrix containing $g_k$ in its columns. The goal is to decouple the overlapping spatiotemporal modes from the network sequence $G$ using -- 1. Graph Dynamic Mode Decomposition algorithm, and 2. a novel Deep Learning-based Graph Dynamic Mode Decomposition algorithm.
\subsection{Graph Dynamic Mode Decomposition} \label{sec:GDMD}
GraphDMD \cite{GraphDMD} assumes that $g_k$ follows an LDS:
\begin{equation} \label{eq_lds}
    g_{k+1} = A g_k
\end{equation}
where $A \in \mathbb{R}^{n^2 \times n^2}$ is a linear operator that shifts the current state $g_k$ to the state at the next time frame $g_{k+1}$. To extract the low dimensional global network dynamics, GraphDMD projects $A$ into a lower dimensional space $\hat{A}$ using tensor-train decomposition, applies eigendecomposition of $\hat{A}$, and projects the eigenvectors back to the original space which we refer to as dynamic modes (DMs). GraphDMD uses tensor-train decomposition to maintain the network structure of $g_k$ and thus, the DMs from GraphDMD can be reshaped into $n\times n$ adjacency matrix forms. Let's assume these DMs are $\Phi_1, \Phi_2, \cdots, \Phi_r$ where $\Phi_p \in \mathbb{C}^{n \times n}$ and the corresponding eigenvalues are $\lambda_1, \lambda_2, \cdots, \lambda_r$ where $\lambda_p \in \mathbb{C}$ (Figure~\ref{fig:fig1}b). Here, $r$ is the total number of DMs. $\Phi_p$ corresponds to the coherent spatial mode and $\lambda_p$ defines its temporal characteristics (growth/decay rate and frequencies). We can see this by unrolling  equation \ref{eq_lds} in time:
\begin{equation} \label{eq:DMD}
    g_{k+1} = A^{k}g_1 = \sum_{p=1}^r \Phi_p \lambda_p^k b_p = \sum_{p=1}^r \Phi_p a_p^k \exp (\omega_p k \Delta t) b_p
\end{equation}
where $\lambda_p = a_p \exp (\omega_p \Delta t)$, $\Phi^\dag$ is the conjugate transpose of $\Phi$, $b_p = vec(\Phi_p^\dag) g_1$ is the projection of the initial value onto the DMD modes, $a_p = ||\lambda_p||$ is the growth/decay rate and $\omega_p = \operatorname{Im}(\ln\lambda_p) / \Delta t$ is the angular frequency of $\Phi_p$.
\subsection{Adaptation of Graph-DMD for Nonlinear Graph Dynamics}
Since the dynamics of the functional networks are often non-linear, the linearity assumption of equation~\ref{eq_lds} is sub-optimal. In this regard, we resort to Koopman operator theory to transform the non-linear system into an LDS using a set of Koopman eigenfunctions $\psi$, i.e., $\psi(g_{k+1}) = A \psi(g_k)$ \cite{Koopman1931HamiltonianSpace}. We learn $\psi$ using a deep autoencoder-based architecture --- DeepGraphDMD --- where the encoder and the decoder are trained to approximate $\psi$ and $\psi^{-1}$, respectively. We enforce $\psi(g_k)$ to follow an LDS by applying Latent Koopman Invariant Loss \cite{lkis} in the form:
\begin{equation}
    \mathcal{L}_{lkis} = ||Y' - (Y'Y^\dag)^{-1}Y||^2_F
\end{equation}
where $ Y = \begin{pmatrix}
\vline & \vline &  & \vline \\
\psi(g_1) & \psi(g_2) & \cdots & \psi(g_{t-s}) \\
\vline & \vline & & \vline
\end{pmatrix}$, $ Y' = \begin{pmatrix}
\vline & \vline &  & \vline \\
\psi(g_2) & \psi(g_3) & \cdots & \psi(g_{t-s + 1}) \\
\vline & \vline &  & \vline
\end{pmatrix}$ are two matrices with columns stacked with $\psi(g_k)$ and $Y^\dag$ is the right inverse of $Y$. After training, we reshape $\psi(g_k)$ into a $n\times n$ network $\psi(G_k)$ and generate the latent network sequence $\psi(G_1), \cdots, \psi(G_{t-s+1})$. We then apply GraphDMD (described in section~\ref{sec:GDMD}) on this latent and linearized network sequence to extract the DMs $\Phi_p$ and their corresponding $\lambda_p$.

However, there are two unique challenges of learning network embeddings using the DeepGraphDMD model: 1. the edge identity and, thus, the interpretability will be lost in the latent space if we directly embed $g_k$ using $\psi$, and 2. $Y^\dag$ doesn't exist, and thus $\mathcal{L}_{lkis}$ can't be computed because $Y$ is low rank with the number of rows $\frac{n (n - 1)}{2} >>$ the number of columns $t - s + 1$.

To solve the first problem, instead of learning $\psi(g_k)$ directly, we embed the BOLD signal $x_k^i$ of each ROI independently using the encoder to learn the latent embeddings $z_k^i$ (Figure~\ref{fig:fig1}a). We then compute the pearson correlation between the latent embeddings of the ROIs to get the Koopman eigenfunctions of $g_k$ i.e., $\psi(g_{k}^{ij}) = \texttt{pearson}(z_k^i, z_k^j)$. The weights of the encoder and decoder are shared across the ROIs.

The second problem arises because the Koopman operator $A$ regresses the value of an edge at the next time-point as a linear combination of all the other edges at the current time-point, i.e., $g^{ij}_{k+1} = \sum_{p,q=1} ^ N w_{pq} g^{pq}_{k}$. This results in $\mathcal{O}(n^2)$ covariates with $t - s + 1 << \mathcal{O}(n^2)$ samples making the regression ill-posed. We propose a sparse Koopman operator where each edge $g^{ij}_k$ is regressed using only the edges that share a common end-point with it, i.e., $g^{ij}_{k+1} = \sum_{p=1, p \neq i,j}^n w_{ip}g^{ip}_{k} + \sum_{q=1, q \neq i,j}^n w_{qj}g^{qj}_k + w_{ij}g^{ij}_k$ (Supplementary Figure~1). Since there are only $\mathcal{O}(n)$ such edges, it solves the ill-posedness of the regression.

% \cite{lkis} shows that under certain assumptions the matrix $Y'Y^\dag$ converges to the linear operator $A$ for which $Y' = A Y$ as $m \to \infty$ and it is also the minimum-norm solution of the linear least-squares regression from the columns of $Y$ to those of $Y'$.

Other than $\mathcal{L}_{lkis}$, we also train the autoencoder with a reconstruction loss $\mathcal{L}_{recon}$ which is the mean-squared error (MSE) between $x_k^i$ and the reconstructed output from the decoder $\hat{x}_k^i$. Moreover, a regularizer $\mathcal{L}_{reg}$ in the form of an MSE loss between $g_k$ and the latent $\psi(g_k)$ is also added. The final loss is the following:
\begin{equation}
    \mathcal{L} = \mathcal{L}_{recon} + \alpha\mathcal{L}_{lkis} + \beta\mathcal{L}_{reg}
\end{equation}
where $\alpha$ and $\beta$ are hyper-parameters. We choose $\alpha$, $\beta$, and other hyperparameters using grid search on the validation set. The network architecture and the values of the hyper-parameters of DeepGraphDMD training are shown in Supplementary Figure 1. The code is available in \footnote{\href{https://github.com/mturja-vf-ic-bd/DeepGraphDMD.git}{https://github.com/mturja-vf-ic-bd/DeepGraphDMD.git}}.
\subsection{Window-based GraphDMD} \label{sec:win_dmd}
We apply GraphDMD in a short window of size 64 time frames with a step size of 4 time frames instead of the whole sequence $G$ because, in real-world fMRI data, both the frequency and the structure of the DMs can change over time. We then combine the DMs across different sliding windows using the following post-processing steps:

\textbf{Post-processing of the DMs:} We first group the DMs within the frequency bins: 0-0.01 Hz, 0.01-0.04Hz, 0.04-0.08Hz, 0.08-0.12Hz, and 0.12-0.16Hz. We then cluster the DMs within each frequency bin using a clustering algorithm and select the cluster centroids as the representative DMs (except for the first bin where we average the DMs). We chose the optimal clustering algorithm to be Spherical KMeans \cite{JMLR:v6:banerjee05a} (among Gaussian Mixture Model, KMeans, Spherical KMeans, DBSCAN, and, KMedoids) and the optimal number of clusters to be 3 for every frequency bin based on silhouette analysis \cite{ROUSSEEUW198753} (Supplementary Figure~2). We use this frequency binning technique to allow for slight variations of $\omega$ of a DM over the scanning session. To align these representative DMs across subjects, we apply another round of spherical clustering on the DMs from all subjects and align them based on their cluster memberships.

\section{Experiments}
\subsection{Dataset} \label{sec:dataset}
% \textbf{Simulated Dataset:}
% We generate a sequence of dynamic adjacency matrices $G$ by first generating 3 time-varying modes $\Phi_1, \Phi_2, \Phi_3$ and sampling 3 corresponding frequencies $\omega_1 \sim \mathcal{N}(0.1, 0.05)$, $\omega_2 \sim \mathcal{N}(1, 0.1)$, $\omega_3 \sim \mathcal{N}(2.5, 0.1)$ (Hz) and plugging them into equation~\ref{eq:DMD}. Each of the $\Phi_p$'s is a $32 \times 32$ block diagonal matrices with block sizes 16, 8, and 4. We choose $a_1=1.01$, $a_2=0.9$, $a_3=1.05$ and $b_1=b_2=b_3=1$. We simulate the process for $k=1, \cdots, 29$ time-points yielding a sequence of 30 matrices of shape $32 \times 32$. We repeat the process ten times with different $\omega_1, \omega_2, \omega_3$ and generate ten matrix sequences.
% \textbf{Human Connectome Project (HCP) Dataset:}
We use rs-fMRI for 840 subjects from the HCP Dense Connectome dataset \footnote{\href{https://www.humanconnectome.org/storage/app/media/documentation/s1200/HCP1200-DenseConnectome+PTN+Appendix-July2017.pdf}{https://www.humanconnectome.org/storage/app/media/documentation/s1200/HCP1200-DenseConnectome+PTN+Appendix-July2017.pdf}} \cite{van2012human}. Each fMRI image was acquired with a temporal resolution ($\Delta t$) of 0.72 s and a 2 mm isotropic spatial resolution using a 3T Siemens Skyra scanner. Individual subjects underwent four rs-fMRI runs of 14.4 min each (1200 frames per run). Group-ICA using FSL's MELODIC tool \cite{hyvarinen1999fast} was applied to parcellate the brain into 50 functional regions (ROIs). To find the correlation between cognition with the rs-fMRI data, we select two behavioral measures related to fluid intelligence: \texttt{CogFluidComp}, \texttt{PMAT24\_A\_CR} and one measure related to crystallized intelligence: \texttt{ReadEng}, and, the normalized scores of the fluid and crystallized cognition measures: \texttt{CogTotalComp}. We regress out the confounding factors: age, gender, and head motion from these behavioral measures using ordinary least squares \cite{liegeois2019resting}.

\subsection{Baseline Methods}
We compare GraphDMD and DeepGraphDMD against three decomposition methods:  Principal Component Analysis (PCA), Independent Component Analysis (ICA), and standard Dynamic Mode Decomposition (DMD) \cite{schmid2010dynamic}. We use the \verb|sklearn| decomposition library for PCA \footnote{sklearn.decomposition.PCA} and ICA \footnote{sklearn.decomposition.FastICA} and the \verb|pyDMD|\footnote{\href{https://mathlab.github.io/PyDMD/dmd.html}{https://mathlab.github.io/PyDMD/dmd.html}} library for standard DMD. We apply PCA and ICA on $g$, and DMD directly on the bold signal $X$ instead of $g$ (for reasons described in section~\ref{sec:result}). We choose the number of components (\verb|n_components|) to be three for these decomposition methods, (Results for other \verb|n_components| values are shown in Supplementary Table~1). The components are aligned across subjects using spherical clustering similar to the GraphDMD modes (Section~\ref{sec:win_dmd}).  We also compare with static functional connectivity (sFC), which is the pairwise pearson correlation between brain regions across all time frames.

\subsection{Simulation Study} \label{sec:sim}
We generate a sequence of dynamic adjacency matrices $G$ using equation~\ref{eq:DMD} from three time-varying modes $\Phi_1, \Phi_2, \Phi_3$ with corresponding frequencies $\omega_1 \sim \mathcal{N}(0.1, 0.05)$, $\omega_2 \sim \mathcal{N}(1, 0.1)$, $\omega_3 \sim \mathcal{N}(2.5, 0.1)$ (Hz). Each $\Phi_p$ is a $32 \times 32$ block diagonal matrices with block sizes 16, 8, and 4. We choose $a_1=1.01$, $a_2=0.9$, $a_3=1.05$ and $b_1=b_2=b_3=1$. We simulate the process for $k=1, \cdots, 29$ time-points yielding a sequence of 30 matrices of shape $32 \times 32$. We repeat the process ten times with different $\omega_1, \omega_2, \omega_3$ and generate ten matrix sequences. We apply PCA, ICA, and GraphDMD (section~\ref{sec:GDMD}) on $G$ to extract three components and compare them against the ground truth modes using pearson correlation.

\subsection{Application of GraphDMD and DeepGraphDMD in HCP data}
\textbf{Comparison of DMs with sFC:}
The ground truth DMs are unknown for the HCP dataset; however, we can use the sFC as a substitute for the ground truth DM with $\omega = 0$ (static DM). sFC offsets the DMs with $\omega > 0$ as they have both positive and negative cycles, and thus only retain the static DM. For comparison, we compute the pearson correlation between the DM within the frequency bin 0-0.01 Hz and sFC for both GraphDMD and DeepGraphDMD. For PCA and ICA, we take the maximum value of the correlation between sFC and the (PCA or ICA) components.

\textbf{Regression Analysis of Behavioral Measures from HCP:}
In this experiment, we regress the behavioral measures with the DMs within each frequency bin (section~\ref{sec:win_dmd}) using Elastic-net. As an input to the Elastic-net, we take the real part of the upper diagonal part of the DM and flatten it into a vector. We then train the Elastic-net in two ways --- 1. single-band: where we train the Elastic-net independently with the DMs in each frequency bin, and 2. multi-band: we concatenate two DMs in the frequency bins: 0-0.01Hz and 0.08-0.12Hz and regress using the concatenated vector. For evaluation, we compute the correlation coefficient $r$ between the predicted and the true values of the measures.

\section{Results}
\subsection{Simulation Study}
\begin{figure}[b]
    \centering
    \includegraphics[width=\textwidth]{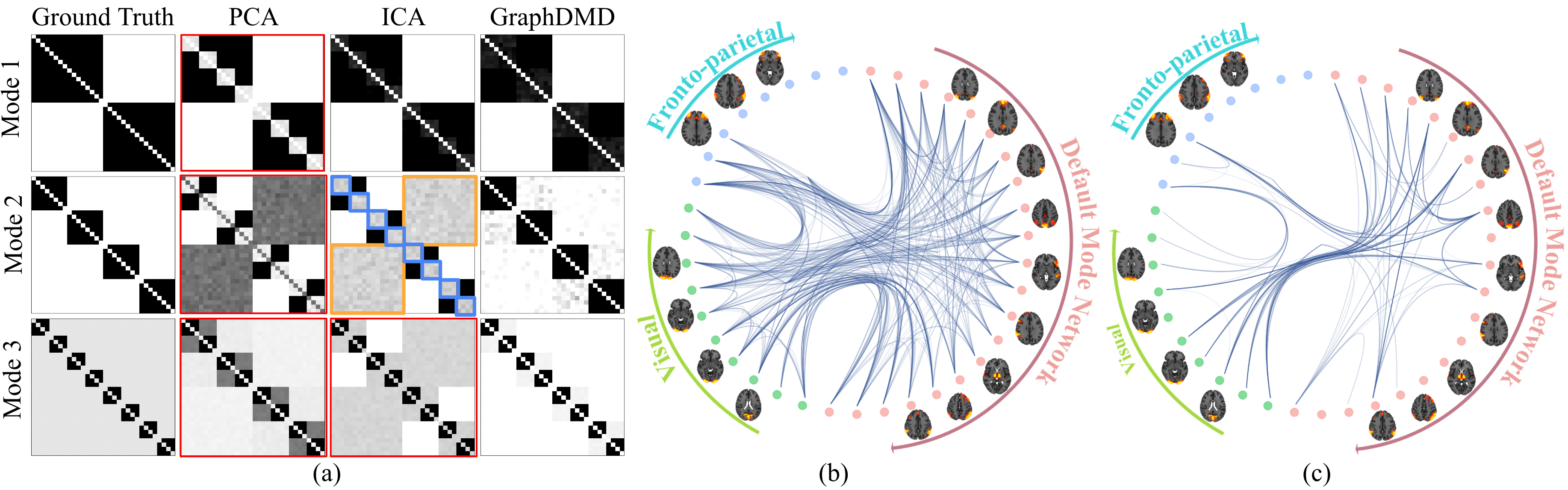}
    \caption{(a) Ground truth network modes from simulated data (column 1) and extracted network modes from PCA (2nd column), ICA (3rd column), and, GraphDMD (4th column), (b) Circle plot of the average DMs with $\omega \approx 0$, (c) $\omega \in [0.08-0.12]$ from DeepGraphDMD organized based on common resting-state networks \cite{SEITZMAN2020116290}.}
    %(d) comparison of the R-squared values for fluid intelligence measure and (e) crystallized intelligence measure.}
    \label{fig:fig2}
\end{figure}

In Figure~\ref{fig:fig2}a, we show the results after applying PCA, ICA, and, GraphDMD on the simulated data described in section~\ref{sec:sim}. Since the DMs in this data are oscillating, the data generated from this process are more likely to be overlapping compared to when the modes are static. As a result, methods that assume static modes, such as PCA and ICA, struggle to decouple the DMs and discover modes in overlapping high-density regions. For example, in Mode 2 of ICA, we can see the remnants of Mode 3 in the blue boxes and Mode 1 (negative cycle) in the orange boxes. We observe similar scenarios in the Mode 3 of ICA and Mode 1, Mode 2, and Mode 3 of PCA (the red boxes). On the other hand, the DMs from GraphDMD have fewer remnants from other modes and closely resemble the ground truth. To empirically compare, the mean ($\pm$ std) pearson correlation for PCA, ICA, and, GraphDMD are $0.81(\pm 0.04)$, $0.88(\pm 0.03)$, and $0.98(\pm 0.01)$.
%Moreover, the GraphDMD also discovers the frequency of each mode to be 0.11, 0.97, and, 2.48 for Mode 1, Mode 2, and Mode 3, respectively. The ability to extract frequencies of the DMs is particularly useful for the interpretability and alignment of the DMs across subjects.

\subsection{Application of GraphDMD and DeepGraphDMD in HCP data} \label{sec:result}
\textbf{Comparison of DMs with sFC:}
The average pearson correlations with sFC across all the subjects are $0.6 (\pm 0.09)$, $0.6 (\pm 0.09)$, $0.84 (\pm 0.09)$, and $0.86 (\pm 0.05)$ for PCA, ICA, GraphDMD, and, DeepGraphDMD (Figure~\ref{fig:fig2}b) respectively. This shows that the DMD-based methods can robustly decouple the static DM from time-varying DMs. In comparison, the corresponding PCA and ICA component has significantly lower correlation due to the overlap from the higher frequency components.

\textbf{Regression Analysis of Behavioral Measures from HCP:}
We show the values of $r$ across different methods in Table~\ref{tab:r_squared}. We only show the results for two frequency bins 0-0.01Hz and 0.08-0.12 Hz, as the DMs in the other bins are not significantly correlated ($r < 0.2$) with the behavioral measures (Supplementary Table~2). The table shows that multi-band training with the DMs from the DMD-based methods significantly improves the regression performance over the baseline methods. Compared to sFC, GraphDMD improves $r$ by 22\%, 6\%, 0.7\%, and, 3\% for \texttt{CogFluidComp}, \texttt{PMAT24\_A\_CR}, \texttt{ReadEng}, \texttt{CogTotalComp}, respectively and DeepGraphDMD further improves the performance by 5\%, 2.2\%, 0.7\%, and, 1.5\%, respectively. Significant performance improvement for \verb|CogFluidComp| can be explained by the DM within the bin 0.08-0.12 Hz. This DM provides additional information related to fluid intelligence ($r = 0.227$ for GraphDMD) to which the sFC doesn't have access. By considering non-linearity, DeepGraphDMD extracts more robust and less noisy DMs (Figure~\ref{fig:fig2}b-c), and hence, it improves the regression performance by $8\%$ compared to GraphDMD in this frequency bin. By contrast, the standard DMD algorithm yields unstable modes with $a_p << 1$ when applied to the network sequence $G$. These modes have no correspondence across subjects and thus can't be used for regression. We instead apply DMD on the BOLD signal $X$, but the DMD modes show little correlation with the behavioral measures. PCA and ICA perform significantly worse than the baseline sFC method for all behavioral measures.

Traditional dynamical functional connectivity analysis methods (such as sliding window-based techniques) consider a sequence of network states. However, our results show that these states can be further decomposed into more atomic network modes. The importance of decoupling these network modes from nonlinearly mixed fMRI signals using DeepGraphDMD has been shown in regressing behavioral measures from HCP data.
\begin{table}[h]
    \centering
    \begin{tabular}{|m{3em}|c|c|c|c|c|}
    \hline
         & Frequency (Hz) & \texttt{CogFluidComp} & \texttt{PMAT24\_A\_CR} & \texttt{ReadEng} & \texttt{CogTotalComp} \\
    \hline
    sFC & N/A & $\mathbf{0.253 \pm 0.003}$ & $0.294 \pm 0.004$ & $0.407 \pm 0.004$ & $\mathbf{0.440 \pm 0.004}$ \\
    \hline
    PCA & N/A & $0.109 \pm 0.003$ & $0.126 \pm 0.003$ & $0.224 \pm 0.003$ & $0.238 \pm 0.003$ \\
    \hline
    ICA & N/A & $0.148 \pm 0.005$ & $0.158 \pm 0.004$ & $0.239 \pm 0.005$ & $0.266 \pm 0.006$ \\
    \hline
    DMD & 0-0.01, 0.08-0.12 & $0.064 \pm 0.002$ & $0.169 \pm 0.002$ & $0.132 \pm 0.003$ & $0.138 \pm 0.006$ \\
    \hline
    \multirow{3}{6em}{Graph \\DMD} & 0-0.01 & $0.254 \pm 0.003$ & $0.289 \pm 0.004$ & $0.402 \pm 0.004$ & $0.438 \pm 0.003$\\ \cline{2-6}
        & 0.08-0.12 & $0.227 \pm 0.004$ & $0.193 \pm 0.004$ & $0.145 \pm 0.004$ & $0.248 \pm 0.004$ \\ \cline{2-6}
         & 0-0.01, 0.08-0.12 & $\mathbf{0.308 \pm 0.004}$ & $0.312 \pm 0.004$ & $0.410 \pm 0.003$ & $\mathbf{0.454 \pm 0.004}$ \\ \cline{2-6}
    \hline
    \multirow{3}{6em}{Deep \\Graph\\ DMD} & 0-0.01 & $0.259 \pm 0.003$ & $0.290 \pm 0.002$ & $0.404 \pm 0.002$ & $0.439 \pm 0.002$\\ \cline{2-6}
        & 0.08-0.12 & $0.245 \pm 0.002$ & $0.201 \pm 0.004$ & $0.144 \pm 0.003$ & $0.251 \pm 0.004$\\ \cline{2-6}
         & 0-0.01, 0.08-0.12 & $\mathbf{0.325 \pm 0.003}$ & $0.319 \pm 0.003$ & $0.413 \pm 0.002$ & $\mathbf{0.461 \pm 0.003}$ \\ \cline{2-6}
    \hline
    \end{tabular}
    \caption{Comparison of $r$ for the behavioral measures across different methods.}
    \label{tab:r_squared}
\end{table}
\section{Conclusion}
In this paper, we proposed a novel algorithm --- DeepGraphDMD --- to decouple spatiotemporal network modes in dynamic functional brain networks. Unlike other decomposition methods, DeepGraphDMD accounts for both the non-linear and the time-varying nature of the functional modes. As a result, these functional modes from DeepGraphDMD are more robust compared to their linear counterpart in GraphDMD and are shown to be correlated with fluid and crystallized intelligence measures.

\bibliographystyle{splncs04}
\bibliography{paper2493}
\end{document}

\title{Supplementary Material --- DeepGraphDMD: Interpretable Spatio-Temporal Decomposition of Non-linear Functional Brain Network Dynamics}
\author{Md Asadullah Turja\inst{1} \and Martin Styner\inst{2} \and Guorong Wu\inst{3}}
%index{Turja, Md Asadullah}
%index{Styner, Martin}
%index{Wu, Guorong}
%
\authorrunning{M.A. Turja et al.}
\institute{Department of Computer Science, University of North Carolina at Chapel Hill,
\email{mturja@cs.unc.edu}\\ \and
Department of Computer Science, University of North Carolina at Chapel Hill,
\email{styner@email.unc.edu}\\ \and
Department of Psychiatry, University of North Carolina at Chapel Hill, \email{guorong\_wu@med.unc.edu}
}
\maketitle              % typeset the header of the contribution

\begin{figure}
\includegraphics[width=\textwidth]{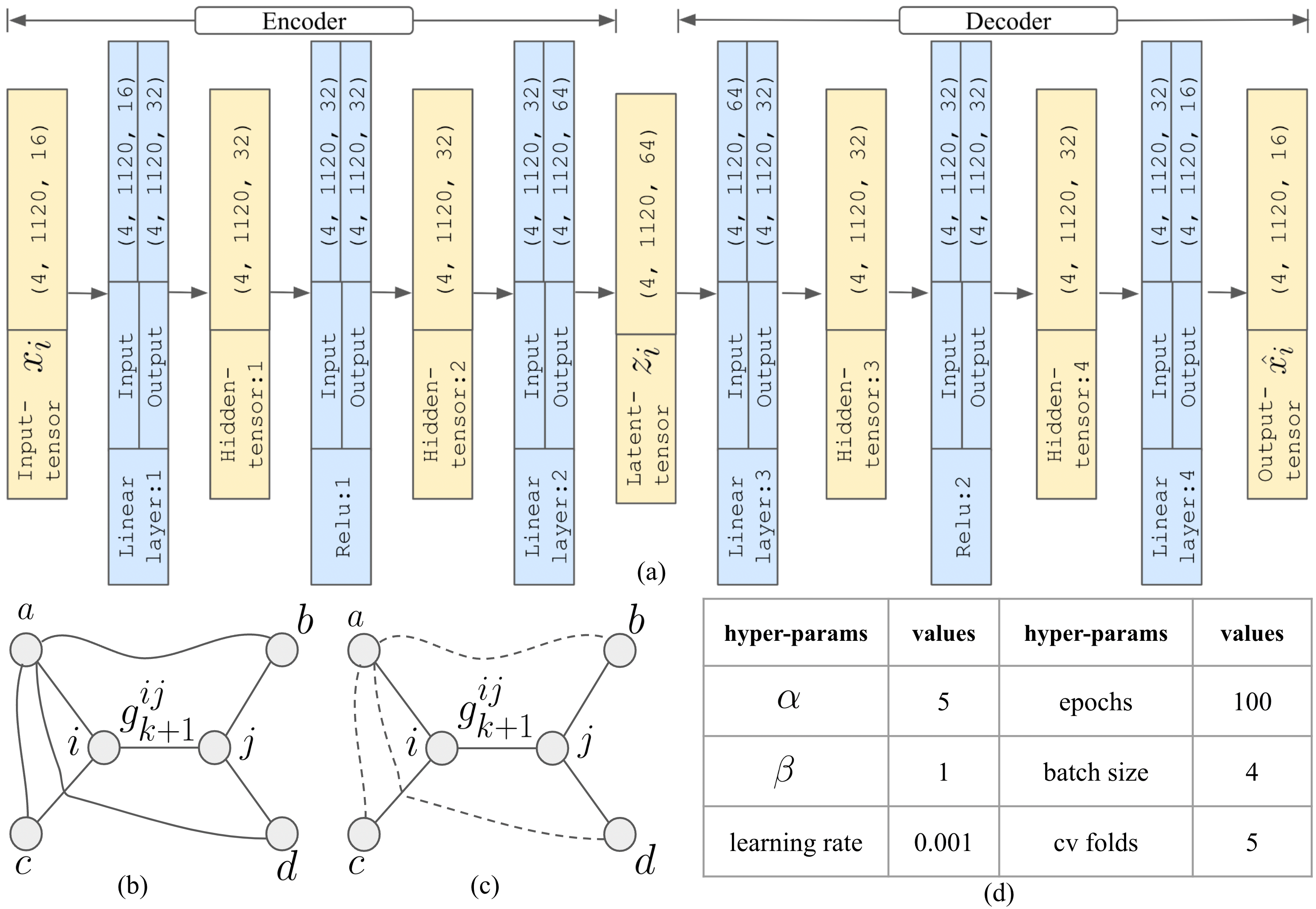}
\caption{(a) The network architecture of DeepGraphDMD, (b) regular Koopman Operator represents $g_{k+1}^{ij}$ as a linear combination of previous values of all the other edges ($g_k^{ia}, g_k^{jb}, g_k^{ab}, g_k^{ic}, g_k^{jd}, g_k^{ad}, g_k^{ij}$) of the graph whereas (c) Sparse Koopman Operator only considers neighboring edges: $g_k^{ia}, g_k^{jb}, g_k^{ic}, g_k^{jd}, g_k^{ij}$ (doesn't consider the dashed edges), and (d) table of hyper-parameters for the DeepGraphDMD training.} \label{fig:supp_fig1}
\end{figure}

\begin{figure}[tb]
\includegraphics[width=\textwidth]{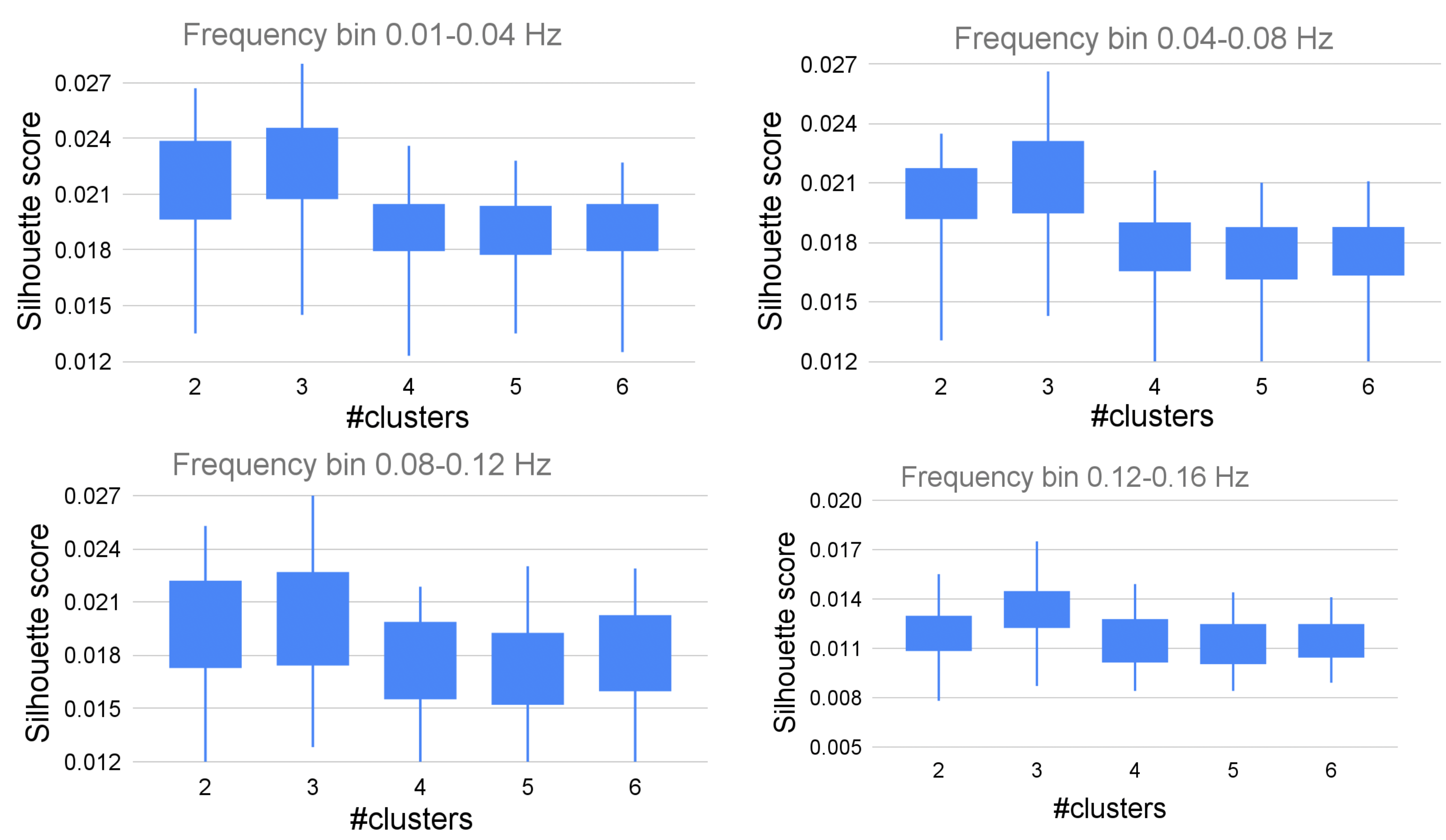}
\caption{Boxplot of the silhouette scores from all the subjects with respect to the number of clusters across different frequency bins (higher is better).}
\label{fig:supp_fig2}
\end{figure}

\begin{table}
    \centering
    \begin{tabular}{|c|c|c|c|c|c|}
    \hline
         \verb|n_components| & 2 & 3 & 4 & 5 & 6 \\
    \hline
         PCA & 0.238 & 0.238 & 0.238 & 0.238 & 0.238\\
    \hline
         ICA & 0.248 & 0.266 & 0.217 & 0.165 & 0.159\\
    \hline
    \end{tabular}
    \caption{The value of $r$ of \texttt{CogTotalComp} for different \texttt{n\_components} in PCA and ICA. $r$ values for PCA don't change because the first PCA component always gives the best $r$ value which doesn't change with \texttt{n\_components}.}
    \label{tab:sup_1}
\end{table}
\begin{table}[tb]
    \centering
    \begin{tabular}{|m{3em}|c|c|c|c|c|}
    \hline
         & Frequency (Hz) & \texttt{CogFluidComp} & \texttt{PMAT24\_A\_CR} & \texttt{ReadEng} & \texttt{CogTotalComp} \\
    \hline
    \multirow{3}{6em}{Graph\\DMD} & 0.01-0.04 & $0.184 \pm 0.012$ & $0.175 \pm 0.009$ & $0.186 \pm 0.017$ & $0.219 \pm 0.021$\\ \cline{2-6}
       & 0.04-0.08 & $0.095 \pm 0.010$ & $0.180 \pm 0.012$& $0.182 \pm 0.021$ & $0.184 \pm 0.021$ \\ \cline{2-6}
         & 0.12-0.16 & $0.156 \pm 0.015$ & $0.167 \pm 0.011$ & $0.183 \pm 0.019$ & $0.192 \pm 0.017$ \\ \cline{2-6}
    \hline
    \multirow{3}{6em}{Deep\\Graph\\DMD} & 0.01-0.04 & $0.181 \pm 0.011$ & $0.182 \pm 0.013$ & $0.182 \pm 0.015$ & $0.207 \pm 0.018$ \\ \cline{2-6}
       & 0.04-0.08 & $0.112 \pm 0.014$ & $0.181 \pm 0.011$ & $0.185 \pm 0.015$ & $0.187 \pm 0.019$ \\ \cline{2-6}
        & 0.12-0.16 & $0.155 \pm 0.011$ & $0.167 \pm 0.016$ & $0.186 \pm 0.012$ & $0.193 \pm 0.014$ \\ \cline{2-6}
    \hline
    \end{tabular}
    \caption{Comparison of $r$ for the behavioral measures for the other frequency bins.}
    \label{tab:sup_2}
\end{table}
%
% the environments 'definition', 'lemma', 'proposition', 'corollary',
% 'remark', and 'example' are defined in the LLNCS documentclass as well.
%
